\documentclass{article}

\usepackage{microtype}
\usepackage{graphicx,array}
\usepackage{subcaption}
\usepackage{booktabs} 
\usepackage{color}
\usepackage{amsmath, amsthm, amssymb, algpseudocode}

\usepackage{hyperref}

\newcommand{\eps}{\varepsilon}
\newcommand{\RR}{\mathbb{R}}
\newcommand{\JPEG}{\mathsf{JPEG}}
\newcommand{\Flow}{\mathsf{Flow}}

\usepackage[arxiv]{icml2019}

\icmltitlerunning{Transfer of Adversarial Robustness Between Perturbation Types}

\newcommand{\minihead}[1]{{\vspace{.25em}\noindent\textbf{#1.} }}

\begin{document}

\twocolumn[
\icmltitle{Transfer of Adversarial Robustness Between Perturbation Types}


\icmlsetsymbol{equal}{*}

\begin{icmlauthorlist}
\icmlauthor{Daniel Kang}{equal,oai,stanford}
\icmlauthor{Yi Sun}{equal,oai,columbia}
\icmlauthor{Tom Brown}{oai}
\icmlauthor{Dan Hendrycks}{berkeley}
\icmlauthor{Jacob Steinhardt}{oai}
\end{icmlauthorlist}

\icmlaffiliation{oai}{OpenAI, San Francisco, CA, USA}
\icmlaffiliation{stanford}{Department of Computer Science, Stanford University, Palo Alto, CA, USA}
\icmlaffiliation{columbia}{Department of Mathematics, Columbia University, New York, NY, USA}
\icmlaffiliation{berkeley}{Department of Electrical Engineering and Computer Science, UC Berkeley, Berkeley, CA, USA}

\icmlcorrespondingauthor{Daniel Kang}{ddkang@stanford.edu}

\vskip 0.3in
]

\printAffiliationsAndNotice{\icmlEqualContribution}

\begin{abstract}
We study the transfer of adversarial robustness of deep neural networks
between different perturbation types.  While most work on
adversarial examples has focused on $L_\infty$ and $L_2$-bounded perturbations, these
do not capture all types of perturbations available to an adversary.  The present work
evaluates 32 attacks of 5 different types against models adversarially trained
on a 100-class subset of ImageNet.  Our empirical results suggest that evaluating
on a wide range of perturbation sizes is necessary to understand whether adversarial
robustness transfers between perturbation types. We further demonstrate that robustness
against one perturbation type \emph{may not always} imply and may sometimes \emph{hurt}
robustness against other perturbation types.  In light of these results, we recommend
evaluation of adversarial defenses take place on a diverse range of perturbation
types and sizes.
\end{abstract}

\section{Introduction}
\label{sec:intro}

Deep networks have shown remarkable accuracy on benchmark
tasks~\cite{he2016identity}, but can also be fooled by
imperceptible changes to inputs, known as adversarial
examples~\cite{goodfellow2014explaining}. In response, researchers have studied
the robustness of models, or how well models generalize in the presence of
(potentially adversarial) bounded perturbations to inputs.

How can we tell if a model is robust? Evaluating model robustness is challenging
because, while evaluating accuracy only requires a fixed distribution,
evaluating the robustness of a model requires
that the model have good performance in the presence of many, potentially hard
to anticipate and model, perturbations. In the context of image classification,
considerable work has focused on robustness to ``$L_{\infty}$-bounded'' perturbations
(perturbations with bounded per-pixel magnitude) \cite{goodfellow2014explaining, madry2017towards,
xie2018feature}. However, models hardened against $L_{\infty}$-bounded perturbations are still vulnerable to
even small, perceptually minor departures from this family, such as small
rotations and translations \cite{engstrom2017rotation}. Meanwhile, researchers continue to develop 
creative attacks that are difficult to even mathematically specify, such as
fake eyeglasses, adversarial stickers, and 3D-printed
objects \cite{sharif2018adversarial, brown2017adversarial, athalye2017synthesizing}.

The perspective of this paper is that any single, simple-to-define type of perturbation 
is likely insufficient to capture what a deployed model will be subject to in the 
real world. To address this, we investigate robustness of models with respect to
a \emph{broad range} of perturbation types. We start with the following question:
\vspace{-0.5em}
\begin{quote}
When and how much does robustness to one type of perturbation transfer to other perturbations?
\end{quote}
\vspace{-0.5em}
We study this question using adversarial training, a strong technique for 
adversarial defense applicable to any fixed attack \cite{goodfellow2014explaining,
madry2017towards}. We evaluate $32$ attacks of $5$ different 
types--$L_\infty$~\cite{goodfellow2014explaining}, $L_2$~\cite{carlini2017towards},
$L_1$~\cite{chen2018ead}, elastic deformations~\cite{xiao2018spatially}, and
JPEG~\cite{shin2017jpeg}--against adversarially trained ResNet-50
models on a 100-class subset of full-resolution ImageNet. 

Our results provide empirical evidence that models robust under one perturbation type
\emph{are not necessarily robust under other natural perturbation types}.  We show that:
\vspace{-0.5em}
\begin{enumerate}
\setlength\itemsep{0.0em}
\item Evaluating on a carefully chosen range of perturbation sizes is important for
measuring robustness transfer.

\item Adversarial training against the elastic deformation attack demonstrates that
adversarial robustness against one perturbation type can transfer poorly to and at times
hurt robustness to other perturbation types.

\item Adversarial training against the $L_2$ attack may be better than training against
the widely used $L_\infty$ attack.
\end{enumerate}
\vspace{-0.5em}
While any given set of perturbation types may not encompass all potential perturbations
that can occur in practice, our results demonstrate that robustness can fail to transfer
even across a small but diverse set of perturbation types.  Prior work in this area
\cite{sharma2017attacking, jordan2019quantifying, tramer2019adversarial} has studied
transfer using single values of $\eps$ for each attack on lower resolution datasets; we
believe our larger-scale study provides a more comprehensive and interpretable view on
transfer between these attacks.  We therefore suggest considering performance against
several different perturbation types and sizes as a first step for rigorous evaluation
of adversarial defenses.

\section{Adversarial attacks} \label{sec:attacks}

\begin{figure}[!ht]
\centering
\begin{tabular}{
>{\centering\arraybackslash}m{.85in}
>{\centering\arraybackslash}m{.85in}
>{\centering\arraybackslash}m{.85in}
}
\includegraphics[width=.85in]{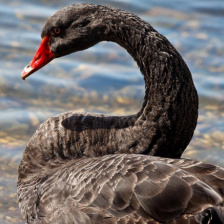} & \includegraphics[width=.85in]{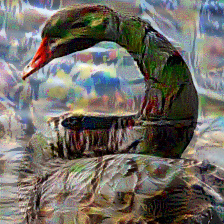} & \includegraphics[width=.85in]{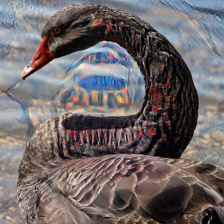}\\
clean & $L_\infty$ & $L_2$\\
\includegraphics[width=.85in]{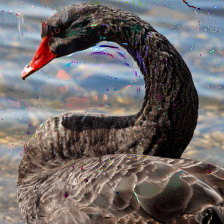} &\includegraphics[width=.85in]{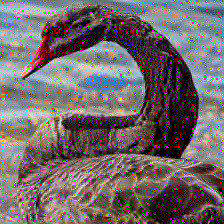} & \includegraphics[width=.85in]{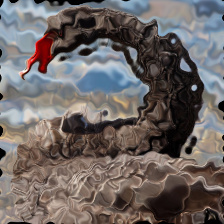} \\
$L_1$ & JPEG & elastic 
\end{tabular} \caption{Sample attacked images with label ``black swan'' for $\eps$ at the top end of our range. \label{fig:sample-images}}
\end{figure}

We consider five types of adversarial attacks under the following framework.
Let $f: \RR^{3 \times 224 \times 224} \to \RR^{100}$ be a model mapping images to
logits\footnote{For all experiments, the input is a $224 \times 224$ image,
and the output is one of $100$ classes.}, and let $\ell(f(x), y)$ denote the
cross-entropy loss. For an input $x$ with true label $y$
and a target class $y' \neq y$, the attacks attempt to find $x'$ such that
\vspace{-0.5em}
\begin{enumerate}
\setlength\itemsep{0.0em}
\item the attacked image $x'$ is a perturbation of $x$, constrained in a sense
which differs for each attack, and

\item[2.] the loss $\ell(f(x'), y')$ is minimized (targeted attack).
\end{enumerate}
\vspace{-0.5em}
We consider the targeted setting and the following attacks, described in more detail below:
\vspace{-0.5em}
\begin{itemize}
\setlength\itemsep{0.0em}
  \item $L_\infty$~\cite{goodfellow2014explaining}
  \item $L_2$~\cite{szegedy2013intriguing,carlini2017towards}
  \item $L_1$~\cite{chen2018ead}
  \item JPEG
  \item Elastic deformation~\cite{xiao2018spatially}
\end{itemize}
\vspace{-0.5em}
The $L_\infty$ and $L_2$ attacks are standard in the adversarial examples
literature~\cite{athalye2018obfuscated, papernot2016distillation, madry2017towards,
carlini2017towards} and we chose the remaining attacks for diversity in
perturbation type. We now describe each attack, with sample images in
Figure \ref{fig:sample-images} and Appendix \ref{sec:attack-samples}.
We clamp output pixel values to $[0, 255]$.

For $L_p$ attacks with $p \in \{1, 2, \infty\}$, the constraint allows an
image $x \in \RR^{3 \times 224 \times 224}$, viewed as a vector of RGB pixel values,
to be modified to an attacked image $x' = x + \delta$ with
\[
\|x' - x\|_p \leq \eps,
\]
where $\|\cdot\|_p$ denotes the $L_p$-norm on $\RR^{3 \times 224 \times 224}$. For the
$L_\infty$ and $L_2$ attacks, we optimize using randomly-initialized projected
gradient descent (PGD), which optimizes the perturbation $\delta$ by gradient
descent and projection to the $L_\infty$ and $L_2$ balls \cite{madry2017towards}.
For the $L_1$ attack, we use the randomly-initialized Frank-Wolfe
algorithm~\cite{frank1956algorithm}, detailed in Appendix \ref{sec:fw-pseudo}.
We believe that our Frank-Wolfe algorithm is more principled than the optimization used in 
existing $L_1$ attacks such as EAD.

As discussed in \citet{shin2017jpeg} as a defense, JPEG compression applies a lossy linear
transformation based on the discrete cosine transform (denoted by $\JPEG$) to image space,
followed by quantization. The JPEG attack, which we believe is new to this work, 
imposes on the attacked image $x'$ an $L_\infty$-constraint in this transformed space:
\[
\|\JPEG(x) - \JPEG(x')\|_\infty \leq \eps.
\]
We optimize $z = \JPEG(x')$ with randomly initialized PGD and apply a right inverse of $\JPEG$
to obtain the attacked image.

The elastic deformation attack allows perturbations
\[
x' = \Flow(x, V),
\]
where $V: \{1, \ldots, 224\}^2 \to \RR^2$ is a vector field on pixel space, and $\Flow$ sets
the value of pixel $(i, j)$ to the (bilinearly interpolated) value at $(i, j) + V(i, j)$.
We constrain $V$ to be the convolution of a vector field $W$ with
a $25 \times 25$ Gaussian kernel with standard deviation $3$, and enforce that
\[
\|W(i, j)\|_\infty \leq \eps \qquad \text{ for } i, j \in \{1, \ldots, 224\}.
\]
We optimize the value of $W$ with randomly initialized PGD.  Note that our attack
differs in details from \citet{xiao2018spatially}, but is similar in spirit.

\begin{figure*}[!ht]
  \includegraphics[width=0.99\linewidth]{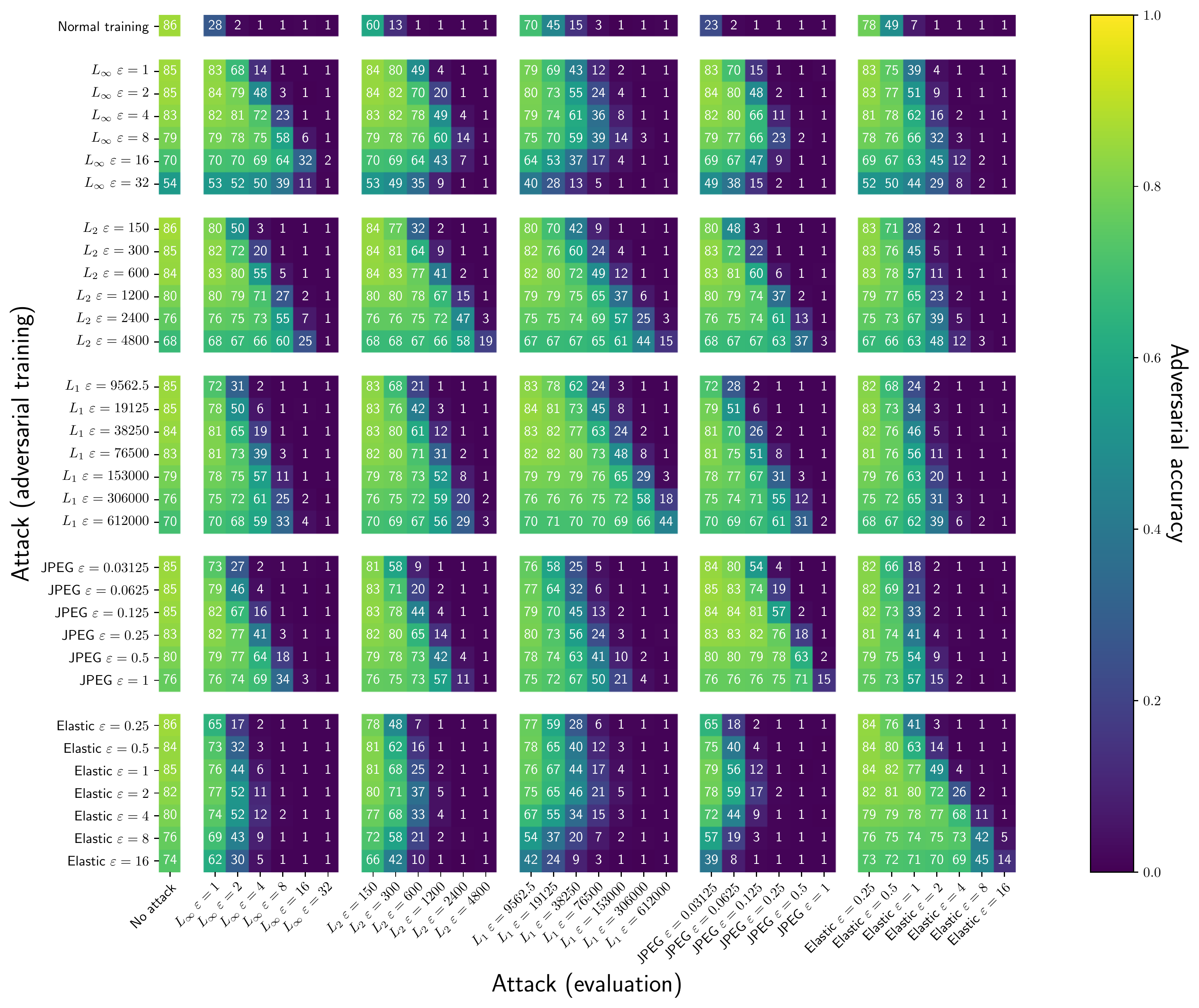}
  \caption{Evaluation accuracies of adversarial attacks (columns) against adversarially
  trained models (rows).\label{fig:grid}}
\end{figure*}

\section{Experiments} \label{sec:experiments}

We measure transfer of adversarial robustness by evaluating our attacks against
adversarially trained models. For each attack, we adversarially train models against
the attack for a range of perturbation sizes $\eps$. We then evaluate each adversarially
trained model against each attack, giving the $2$-dimensional accuracy grid of attacks
evaluated against adversarially trained models shown in Figure \ref{fig:grid} (analyzed
in detail in Section~\ref{sec:results}).

\subsection{Experimental setup}

\minihead{Dataset and model}
We use the $100$-class subset of ImageNet-1K~\cite{deng2009imagenet} containing classes whose
WordNet ID is a multiple of $10$.  We use the ResNet-50~\cite{he2016identity} architecture with 
standard 224$\times$224 resolution as implemented in \texttt{torchvision}.
We believe this full resolution is necessary for the elastic and JPEG attacks.

\minihead{Training hyperparameters}
We trained on machines with 8 Nvidia V100 GPUs using standard data augmentation
practices~\cite{he2016identity}.  Following best practices for multi-GPU training~\cite{goyal2017accurate},
we used synchronized SGD for $90$ epochs with a batch size of 32$\times$8 and a learning rate schedule
in which the learning rate is ``warmed up'' for 5 epochs and decayed at epochs 30, 60, and 80
by a factor of 10. Our initial learning rate after warm-up was 0.1, momentum was $0.9$,
and weight decay was $5\times10^{-6}$.

\begin{table*}[ht!] \centering
\begin{tabular}{lllllll} \hline
attack & optimization algorithm & $\eps$ or $\eps_{\text{max}}$ values & step size & steps (adversarial training) & steps (eval)\\ \hline
$L_\infty$ & PGD  & $\{2^i \mid 0 \leq i \leq 5\}$ & $\frac{\eps}{\sqrt{\text{steps}}}$ & 10 & 50\\
$L_2$      & PGD  & $\{150 \cdot 2^i \mid 0 \leq i \leq 5\}$ &  $\frac{\eps}{\sqrt{\text{steps}}}$ & 10 & 50\\
$L_1$      & Frank-Wolfe & $\{9562.5 \cdot 2^i \mid 0 \leq i \leq 6\}$ & N/A & 10 & 50\\
JPEG       & PGD & $\{0.03125 \cdot 2^i \mid 0 \leq i \leq 5\}$ & $\frac{\eps}{\sqrt{\text{steps}}}$ & 10 & 50\\
Elastic    & PGD & $\{0.25 \cdot 2^i \mid 0 \leq i \leq 6\}$ & $\frac{\eps}{\sqrt{\text{steps}}}$ & 30 & 100\\\hline
\end{tabular} \caption{Attack parameters for adversarial training and evaluation \label{tab:adv-settings}}
\end{table*}

\minihead{Adversarial training}
We harden models against attacks using adversarial training~\cite{madry2017towards}.
To train against attack $A$, for each mini-batch of training images,
we select target classes for each image uniformly at random from the $99$ incorrect classes.
We generate adversarial images by applying the targeted attack $A$ to the current model
with $\eps$ chosen uniformly at random between $0$ and $\eps_{\text{max}}$.
Finally, we update the model with a step of synchronized SGD using these adversarial images alone.

We list attack parameters used for training in Table \ref{tab:adv-settings}. For the PGD attack,
we chose step size $\frac{\eps}{\sqrt{\text{steps}}}$, motivated by the
fact that taking step size proportional to $1/\sqrt{\text{steps}}$ is optimal
for non-smooth convex functions \citep{nemirovski1978cezari,nemirovski1983complexity}. 
Note that the greater number of PGD steps for elastic deformation is due to the
greater difficulty of its optimization problem, which we are not confident is
fully solved even with this greater number of steps.

\minihead{Attack hyperparameters}
We evaluate our adversarially trained models on the (subsetted) ImageNet-1K validation
set against targeted attacks with target chosen uniformly at random from among
the $99$ incorrect classes.  We list attack parameters for evaluation in Table
\ref{tab:adv-settings}. As suggested in \cite{carlini2019evaluating}, we use more steps
for evaluation than for adversarial training to ensure PGD converges.

\subsection{Results and analysis} \label{sec:results}

Using the results of our adversarial training and evaluation experiments in
Figure~\ref{fig:grid}, we draw the following conclusions.

\minihead{Choosing $\eps$ well is important} Because attack strength increases
with the allowed perturbation magnitude $\eps$, comparing robustness between
different perturbation types requires a careful choice of $\eps$ for both attacks.
First, we observe that a \emph{range} of $\eps$ yielding comparable attack strengths
should be used for all attacks to avoid drawing misleading conclusions.  We suggest the
following principles for choosing this range, which we followed for the parameters in
Table \ref{tab:adv-settings}:
\vspace{-0.75em}
\begin{enumerate}
\setlength\itemsep{0.0em}
  \item Models adversarially trained against the minimum value of $\eps$ should
  have validation accuracy comparable to that of a model trained on unattacked data.

  \item Attacks with the maximum value of $\eps$ should substantially reduce
  validation accuracy in adversarial training or perturb the images
  enough to confuse humans.
\end{enumerate}
\vspace{-0.75em}
To illustrate this point, we provide in Appendix \ref{sec:trunc-range} a subset of
Figure \ref{fig:grid} with $\eps$ ranges that differ in strength between attacks;
the (deliberately) biased ranges of $\eps$ chosen in this subset cause the $L_1$
and elastic attacks to be perceived as stronger than our full results reveal.

Second, even if two attacks are evaluated on ranges of $\eps$ of comparable
strength, the specific values of $\eps$ chosen within those ranges may be important.
In our experiments, we scaled $\eps$ geometrically for all attacks, but when interpreting
our results, attack strength may not scale in the same way
with $\eps$ for different attacks. As a result, we only draw conclusions which are
invariant to the precise scaling of attack strength with $\eps$. We illustrate
this type of analysis with the following two examples.

\minihead{Robustness against elastic transfers poorly to the other attacks}
In Figure \ref{fig:grid}, the accuracies of models adversarially trained against
elastic are higher against elastic than the other attacks, meaning that for these
values of $\eps$, robustness against elastic does not imply robustness against other
attacks.  On the other hand, training against elastic with $\eps \geq 4$ generally
increases accuracy against elastic with $\eps \geq 4$, but decreases accuracy
against all other attacks.

Together, these imply that the lack of transfer we observe in Figure \ref{fig:grid} is
not an artifact of the specific values of $\eps$ we chose, 
but rather a broader effect at the level of perturbation types.
In addition, this example shows that increasing robustness to larger perturbation sizes of a given type
can \emph{hurt} robustness to other perturbation types.  This effect
is only visible by considering an appropriate range of $\eps$ and cannot be 
detected from a single value of $\eps$ alone.

\minihead{$L_2$ adversarial training is weakly better than $L_\infty$}
Comparing rows of Figure \ref{fig:grid} corresponding to training against $L_2$
with $\eps \in \{300, 600, 1200, 2400, 4800\}$ with rows corresponding to training
against $L_\infty$ with $\eps \in \{1, 2, 4, 8, 16\}$, we see that training against $L_2$
yields slightly lower accuracies against $L_\infty$ attacks and higher accuracies
against all other attacks.  Because this effect extends to all $\eps$ for which
training against $L_\infty$ is helpful, it does not depend on the relation between
$L_\infty$ attack strength and $\eps$.  In fact, against the stronger half of our attacks,
training against $L_2$ with $\eps = 4800$ gives comparable or better accuracy to
training against $L_\infty$ with adaptive choice of $\eps$.  This provides some
evidence that $L_2$ is more effective to train against than $L_\infty$.

\section{Conclusion}
\label{sec:conclusion}

This work presents an empirical study of when and how much robustness transfers
between different adversarial perturbation types.  Our results on adversarial training
and evaluation of 32 different attacks on a 100-class subset of ImageNet-1K highlight
the importance of considering a diverse range of perturbation sizes and types for
assessing transfer between types, and we recommend this as a guideline for evaluating
adversarial robustness.

\section*{Acknowledgements}

D.~K.~was supported by NSF Grant DGE-1656518. Y.~S.~was supported by a Junior Fellow award from the Simons Foundation and NSF Grant DMS-1701654. D.~K., Y.~S., and J.~S.~were supported by a grant from the Open Philanthropy Project.

\bibliography{adv-icml2019}
\bibliographystyle{icml2019}

\appendix

\section{Sample attacked images} \label{sec:attack-samples}

\begin{figure*}[!ht]\centering
\begin{tabular}{
>{\centering\arraybackslash}m{.4in}
>{\centering\arraybackslash}m{.85in}
>{\centering\arraybackslash}m{.85in}
>{\centering\arraybackslash}m{.85in}
>{\centering\arraybackslash}m{.85in}
>{\centering\arraybackslash}m{.85in}
>{\centering\arraybackslash}m{.85in}}
& clean & $L_\infty$ & $L_2$ & $L_1$  & JPEG & elastic \\
& & $\eps=32$ & $\eps=4800$ & $\eps=306000$  & $\eps=1$& $\eps=8$ \\
black swan & \includegraphics[width=.85in]{figures/norm_rand/clean_10} & \includegraphics[width=.85in]{figures/norm_rand/linf_32_10} & \includegraphics[width=.85in]{figures/norm_rand/l2_4800_10} & \includegraphics[width=.85in]{figures/norm_rand/l1_2_10} & \includegraphics[width=.85in]{figures/norm_rand/jpeg_1_10} & \includegraphics[width=.85in]{figures/norm_rand/elastic_8_10}\\
chain mail & \includegraphics[width=.85in]{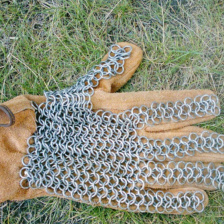} & \includegraphics[width=.85in]{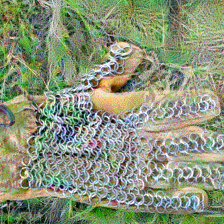} & \includegraphics[width=.85in]{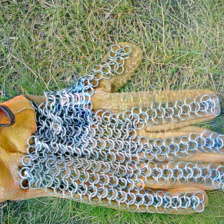} &\includegraphics[width=.85in]{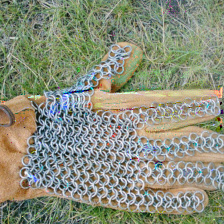}  & \includegraphics[width=.85in]{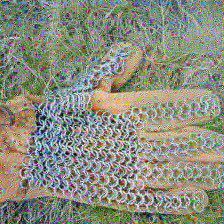}& \includegraphics[width=.85in]{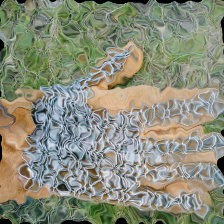} \\
espresso maker & \includegraphics[width=.85in]{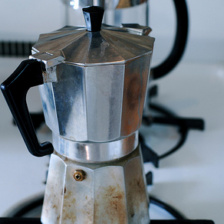} & \includegraphics[width=.85in]{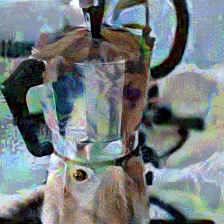} & \includegraphics[width=.85in]{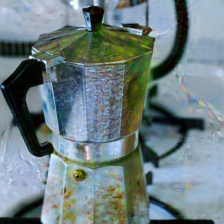} &\includegraphics[width=.85in]{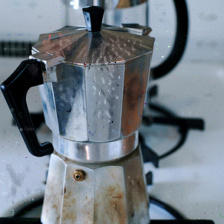}  & \includegraphics[width=.85in]{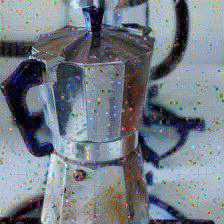}& \includegraphics[width=.85in]{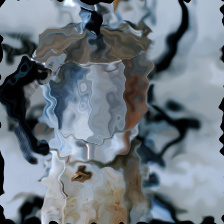}\\
manhole cover & \includegraphics[width=.85in]{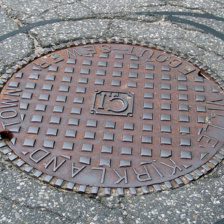} & \includegraphics[width=.85in]{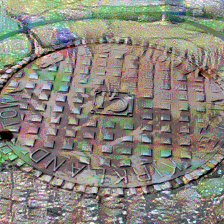} & \includegraphics[width=.85in]{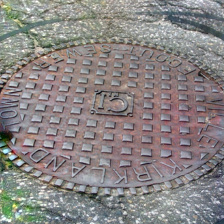} &\includegraphics[width=.85in]{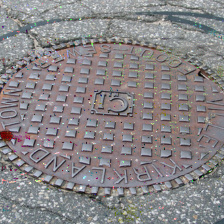}  & \includegraphics[width=.85in]{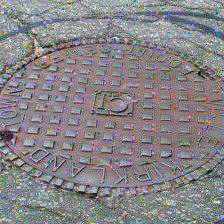}& \includegraphics[width=.85in]{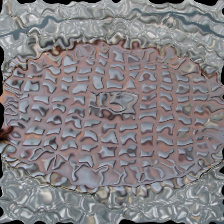} \\
water tower &\includegraphics[width=.85in]{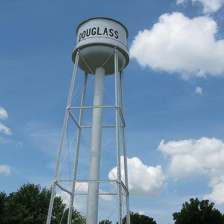} & \includegraphics[width=.85in]{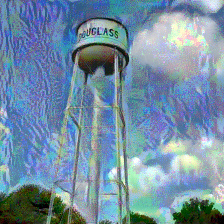} & \includegraphics[width=.85in]{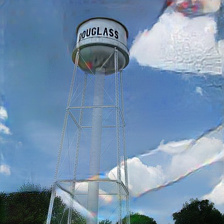} &\includegraphics[width=.85in]{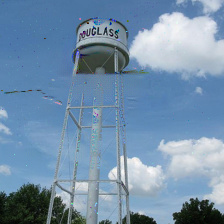} & \includegraphics[width=.85in]{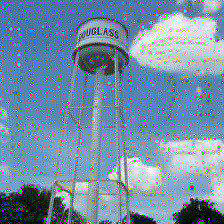} & \includegraphics[width=.85in]{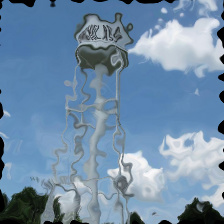} \\
orange & \includegraphics[width=.85in]{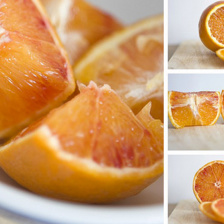} & \includegraphics[width=.85in]{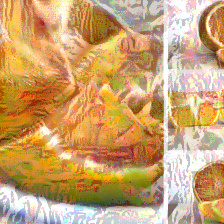} & \includegraphics[width=.85in]{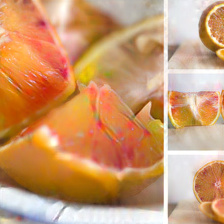} &\includegraphics[width=.85in]{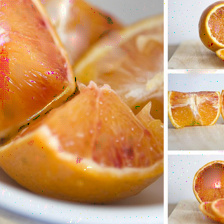}  & \includegraphics[width=.85in]{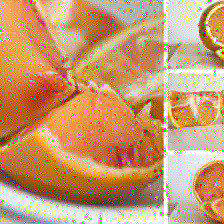}& \includegraphics[width=.85in]{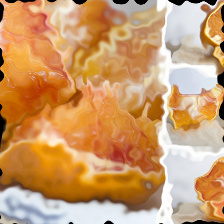} \\
volcano & \includegraphics[width=.85in]{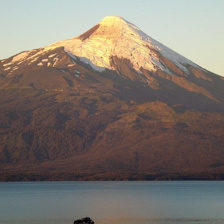} & \includegraphics[width=.85in]{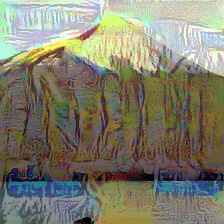} & \includegraphics[width=.85in]{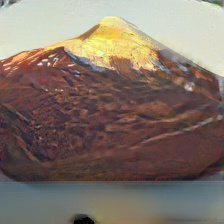} &\includegraphics[width=.85in]{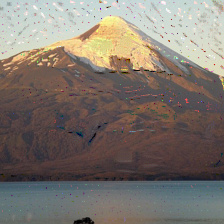} & \includegraphics[width=.85in]{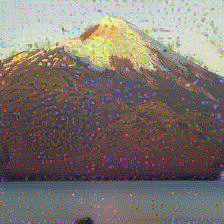} & \includegraphics[width=.85in]{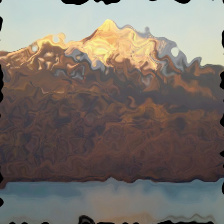} \\
\end{tabular} \caption{Strong attacks applied to sample images\label{fig:strong-attack}}
\end{figure*}

\begin{figure*}[!ht]\centering
\begin{tabular}{
>{\centering\arraybackslash}m{.4in}
>{\centering\arraybackslash}m{.85in}
>{\centering\arraybackslash}m{.85in}
>{\centering\arraybackslash}m{.85in}
>{\centering\arraybackslash}m{.85in}
>{\centering\arraybackslash}m{.85in}
>{\centering\arraybackslash}m{.85in}}
& clean & $L_\infty$ & $L_2$  & $L_1$  & JPEG& elastic\\
& & $\eps=4$ & $\eps=600$& $\eps=38250$  & $\eps=0.125$ & $\eps=1$ \\
black swan & \includegraphics[width=.85in]{figures/norm_rand/clean_10} & \includegraphics[width=.85in]{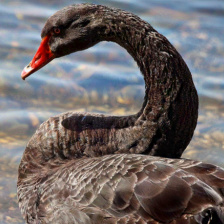} & \includegraphics[width=.85in]{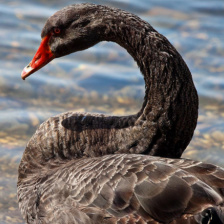} &\includegraphics[width=.85in]{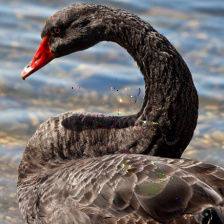} & \includegraphics[width=.85in]{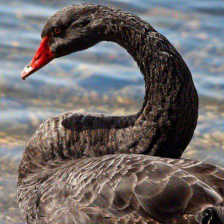} & \includegraphics[width=.85in]{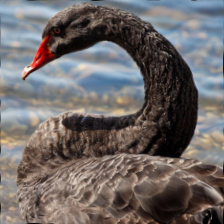}\\
chain mail & \includegraphics[width=.85in]{figures/norm_rand/clean_49} & \includegraphics[width=.85in]{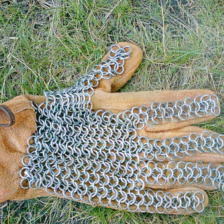} & \includegraphics[width=.85in]{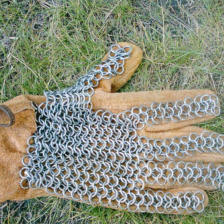} &\includegraphics[width=.85in]{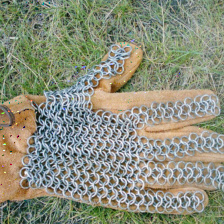}  & \includegraphics[width=.85in]{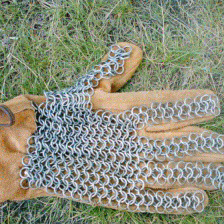}& \includegraphics[width=.85in]{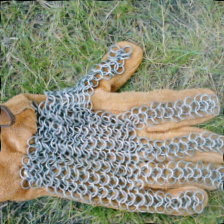}\\
espresso maker & \includegraphics[width=.85in]{figures/norm_rand/clean_55} & \includegraphics[width=.85in]{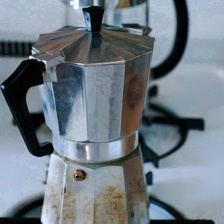} & \includegraphics[width=.85in]{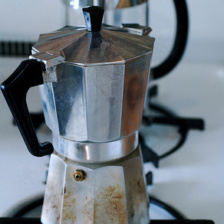} &\includegraphics[width=.85in]{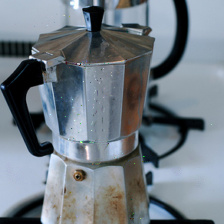}  & \includegraphics[width=.85in]{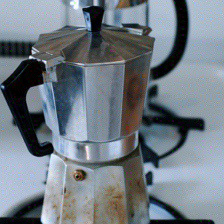}& \includegraphics[width=.85in]{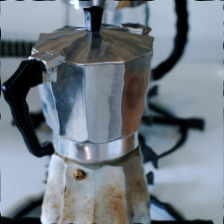}\\
manhole cover & \includegraphics[width=.85in]{figures/norm_rand/clean_64} & \includegraphics[width=.85in]{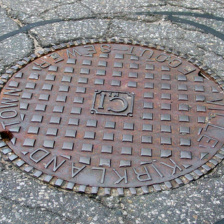} & \includegraphics[width=.85in]{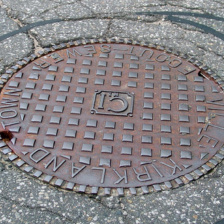} &\includegraphics[width=.85in]{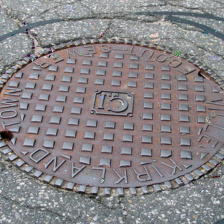} & \includegraphics[width=.85in]{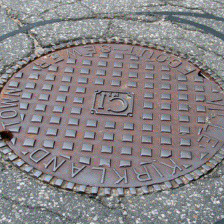} & \includegraphics[width=.85in]{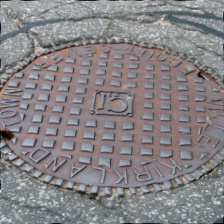}\\
water tower &\includegraphics[width=.85in]{figures/norm_rand/clean_90} & \includegraphics[width=.85in]{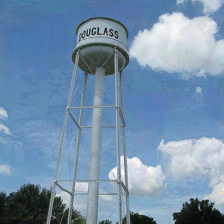} & \includegraphics[width=.85in]{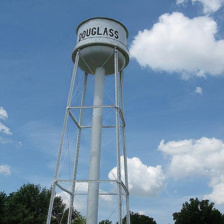} &\includegraphics[width=.85in]{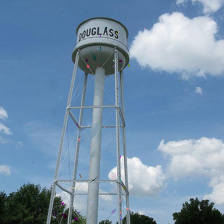} & \includegraphics[width=.85in]{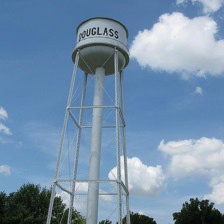} & \includegraphics[width=.85in]{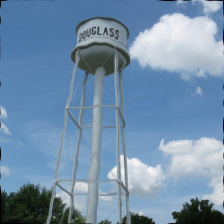}\\
orange & \includegraphics[width=.85in]{figures/norm_rand/clean_95} & \includegraphics[width=.85in]{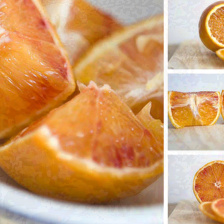} & \includegraphics[width=.85in]{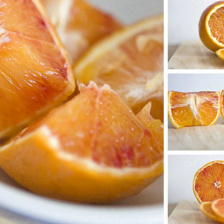} &\includegraphics[width=.85in]{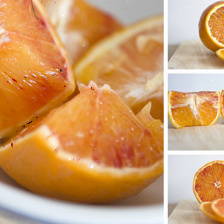} & \includegraphics[width=.85in]{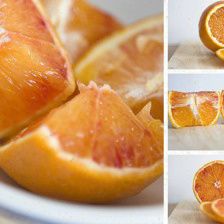} & \includegraphics[width=.85in]{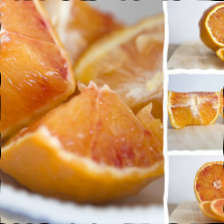}\\
volcano & \includegraphics[width=.85in]{figures/norm_rand/clean_98} & \includegraphics[width=.85in]{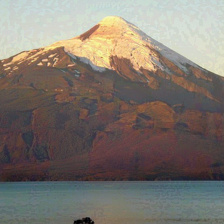} & \includegraphics[width=.85in]{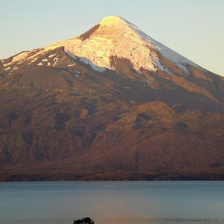} &\includegraphics[width=.85in]{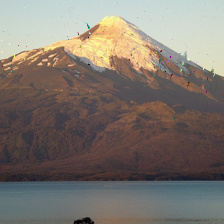} & \includegraphics[width=.85in]{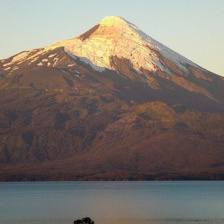} & \includegraphics[width=.85in]{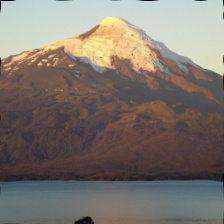}\\
\end{tabular} \caption{Weak attacks applied to sample images\label{fig:weak-attack}}
\end{figure*}

\begin{figure*}[!ht]\centering
\begin{tabular}{
>{\centering\arraybackslash}m{.6in}
>{\centering\arraybackslash}m{.7in}
>{\centering\arraybackslash}m{.7in}
>{\centering\arraybackslash}m{.7in}
>{\centering\arraybackslash}m{.7in}
>{\centering\arraybackslash}m{.7in}
>{\centering\arraybackslash}m{.7in}
>{\centering\arraybackslash}m{.7in}}
attack & clean  & $L_2$ $\eps=2400$ & $L_2$ $\eps=2400$ & $L_1$ $\eps = 153000$ & $L_1$ $\eps=153000$ & elastic $\eps=4$ & elastic $\eps=4$ \\
adversarial training& & $L_1$ $\eps = 153000$ & elastic $\eps = 4$ & $L_2$ $\eps=2400$& elastic $\eps=4$ &  $L_2$ $\eps=2400$ & $L_1$ $\eps=153000$\\
black swan & \includegraphics[width=.7in]{figures/norm_rand/clean_10} & \includegraphics[width=.7in]{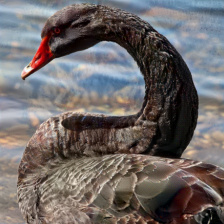} & \includegraphics[width=.7in]{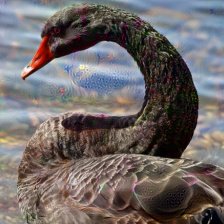} & \includegraphics[width=.7in]{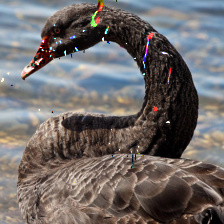} & \includegraphics[width=.7in]{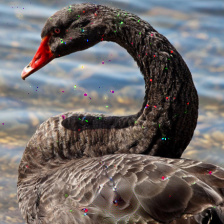}& \includegraphics[width=.7in]{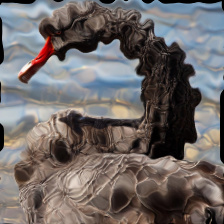} & \includegraphics[width=.7in]{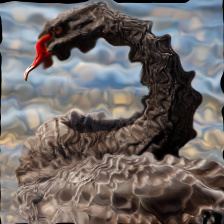}  \\
chain mail & \includegraphics[width=.7in]{figures/norm_rand/clean_49} & \includegraphics[width=.7in]{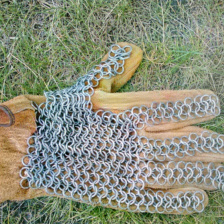} & \includegraphics[width=.7in]{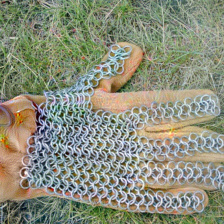} & \includegraphics[width=.7in]{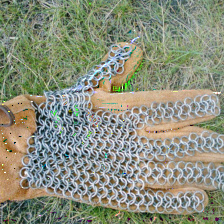} & \includegraphics[width=.7in]{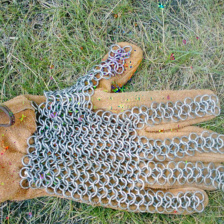}& \includegraphics[width=.7in]{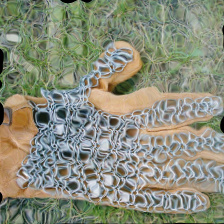} & \includegraphics[width=.7in]{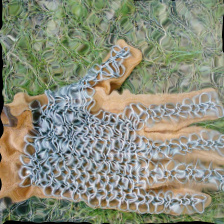}  \\
espresso maker & \includegraphics[width=.7in]{figures/norm_rand/clean_55} & \includegraphics[width=.7in]{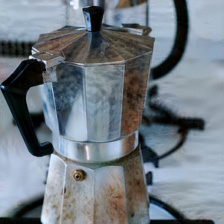} & \includegraphics[width=.7in]{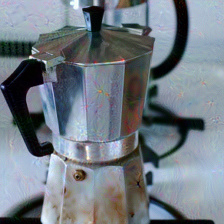} & \includegraphics[width=.7in]{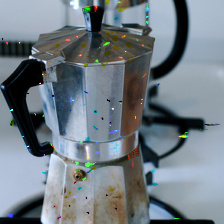} & \includegraphics[width=.7in]{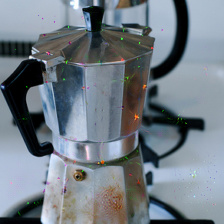}& \includegraphics[width=.7in]{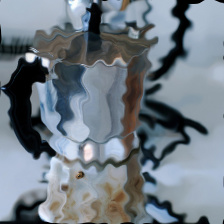} & \includegraphics[width=.7in]{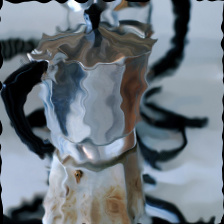}  \\
manhole cover & \includegraphics[width=.7in]{figures/norm_rand/clean_64} & \includegraphics[width=.7in]{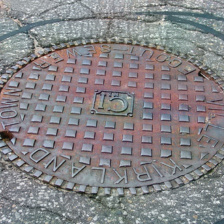} & \includegraphics[width=.7in]{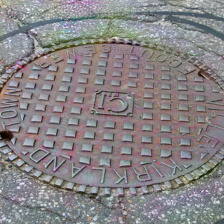} & \includegraphics[width=.7in]{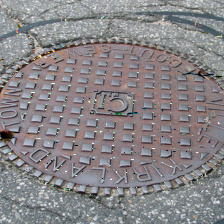} & \includegraphics[width=.7in]{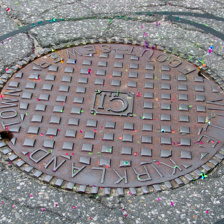}& \includegraphics[width=.7in]{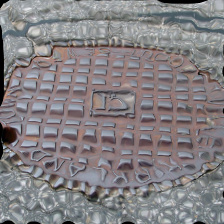} & \includegraphics[width=.7in]{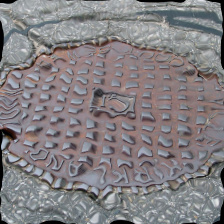}  \\
water tower & \includegraphics[width=.7in]{figures/norm_rand/clean_90} & \includegraphics[width=.7in]{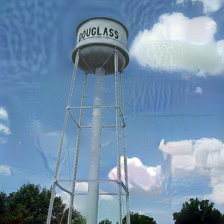} & \includegraphics[width=.7in]{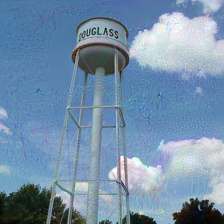} & \includegraphics[width=.7in]{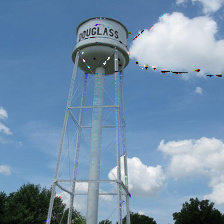} & \includegraphics[width=.7in]{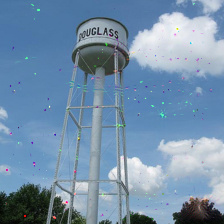}& \includegraphics[width=.7in]{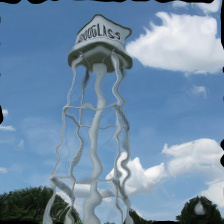} & \includegraphics[width=.7in]{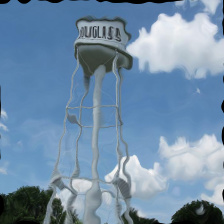}  \\
orange & \includegraphics[width=.7in]{figures/norm_rand/clean_95} & \includegraphics[width=.7in]{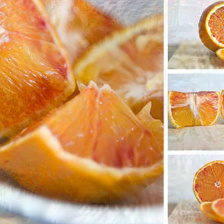} & \includegraphics[width=.7in]{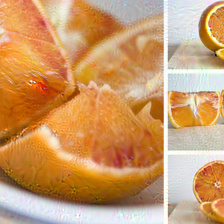} & \includegraphics[width=.7in]{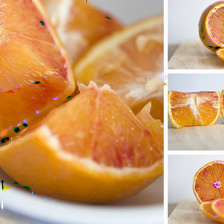} & \includegraphics[width=.7in]{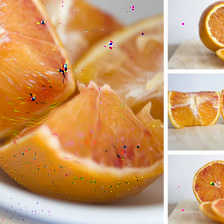}& \includegraphics[width=.7in]{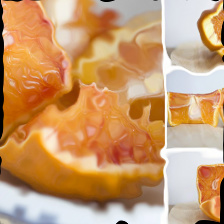} & \includegraphics[width=.7in]{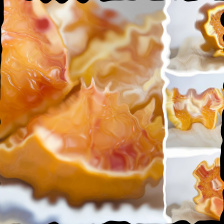}  \\
volcano & \includegraphics[width=.7in]{figures/norm_rand/clean_98} & \includegraphics[width=.7in]{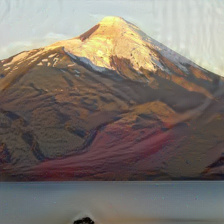} & \includegraphics[width=.7in]{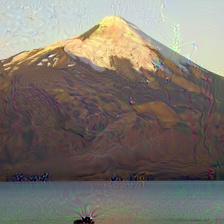} & \includegraphics[width=.7in]{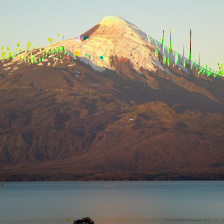} & \includegraphics[width=.7in]{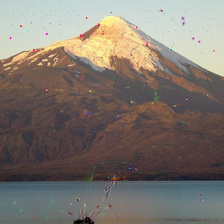}& \includegraphics[width=.7in]{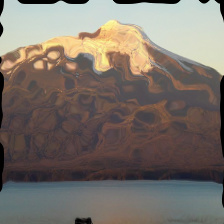} & \includegraphics[width=.7in]{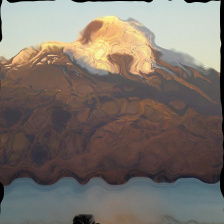}  \\
\end{tabular}\caption{Transfer across attack types \label{fig:attack-transfer}}
\end{figure*}

In this appendix, we give more comprehensive sample outputs for our adversarial attacks.  Figures \ref{fig:strong-attack} and \ref{fig:weak-attack} show sample attacked images for attacks with relatively large and small $\eps$ in our range, respectively.  Figure \ref{fig:attack-transfer} shows examples of how attacked images can be influenced by different types of adversarial training for defense models.  In all cases, the images were generated by running the specified attack against an adversarially trained model with parameters specified in Table \ref{tab:adv-settings} for both evaluation and adversarial training.

\section{Evaluation on a truncated $\eps$ range} \label{sec:trunc-range}

\begin{figure*}[!ht]
  \includegraphics[width=0.99\linewidth]{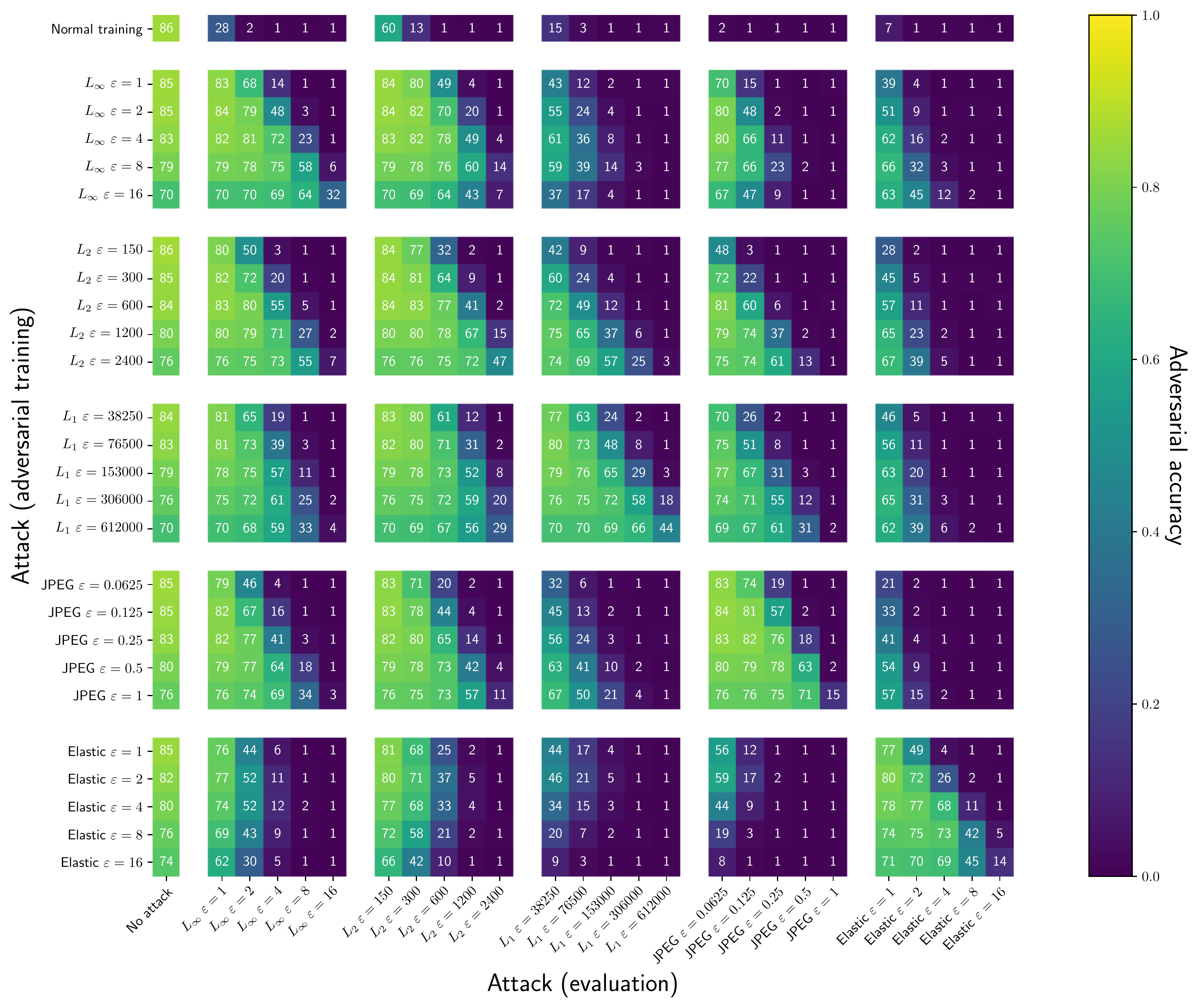}
  \caption{Evaluation accuracies of adversarial attacks (columns) against adversarially
  trained models (rows) for a truncated $\eps$ range.\label{fig:grid-small}}
\end{figure*}

In this appendix, we show in Figure \ref{fig:grid-small} a subset of Figure
\ref{fig:grid} with a truncated range of $\eps$.  In particular, we omitted
small values of $\eps$ for $L_1$, elastic, and JPEG and large values of
$\eps$ for $L_\infty$ and $L_2$.  The resulting accuracy grid gives several
misleading impressions, including:
\begin{enumerate}
\item The $L_1$ attack is stronger than $L_\infty$, $L_2$, and JPEG.

\item Training against the other attacks gives almost no robustness against
the elastic attack.
\end{enumerate}
The full range of results in Figure \ref{fig:grid} shows that these two
purported effects are artifacts of the incorrectly truncated range of $\eps$ used
in Figure \ref{fig:grid-small}.  In particular:
\begin{enumerate}
\item The additional smaller $\eps$ columns for the $L_1$ attack in Figure \ref{fig:grid}
demonstrate its perceived strength in Figure \ref{fig:grid-small} is an artifact of
incorrectly omitting these values.

\item The additional smaller $\eps$ columns for the elastic attack in Figure \ref{fig:grid}
reveal that training against the other attacks is effective in defending against weak
versions of the elastic attack, contrary to the impression given by Figure \ref{fig:grid-small}.
\end{enumerate}

\begin{algorithm*}[!ht]
\caption{Pseudocode for the Frank-Wolfe algorithm for the $L_1$ attack. \label{alg:fw-alg}}
\begin{algorithmic}[1]
\State \textbf{Input:} function $f$, initial input $x \in [0,1]^d$, $L_1$ radius $\rho$, number of steps $T$.
\State \textbf{Output:} approximate maximizer $\bar{x}$ of $f$ over the truncated $L_1$ ball $B_1(\rho; x) \cap [0,1]^d$ centered at $x$.
\State 
\State $x^{(0)} \gets \mathrm{RandomInit}(x)$ \Comment{Random initialization}
\For{$t = 1, \ldots, T$}
  \State $g \gets \nabla f(x^{(t-1)})$ \Comment{Obtain gradient}
  \For{$k = 1, \ldots, d$}
  \State $s_k \gets \text{index of the coordinate of $g$ by with $k^\text{th}$ largest norm}$
  \EndFor
  \State $S_k \gets \{s_1, \ldots, s_k\}$.
  \State
  \For{$i = 1, \ldots, d$} \Comment{Compute move to boundary of $[0, 1]$ for each coordinate.}
  \If{$g_i > 0$}
  \State $b_i \gets 1-x_i$
  \Else
  \State $b_i \gets -x_i$
  \EndIf
  \EndFor

  \State $M_k \gets \sum_{i \in S_k} |b_i|$ \Comment{Compute $L_1$-perturbation of moving $k$ largest coordinates.}
  \State $k^* \gets \max\{k \mid M_k \leq \rho\}$  \Comment{Choose largest $k$ satisfying $L_1$ constraint.}
  \For{$i = 1, \ldots, d$}   \Comment{Compute $\hat{x}$ maximizing $g^{\top}x$ over the $L_1$ ball.}
  \If{$i \in S_{k^*}$}
  \State $\hat{x}_i \gets x_i + b_i$
  \ElsIf{$i = s_{k^* + 1}$}
  \State $\hat{x}_i \gets x_i + (\rho - M_{k^*}) \operatorname{sign}(g_i)$
  \Else
  \State $\hat{x}_i \gets x_i$
  \EndIf
  \EndFor
  \State $x^{(t)} \gets (1-\frac{1}{t})x^{(t-1)} + \frac{1}{t}\hat{x}$ \Comment{Average $\hat{x}$ with previous iterates}
\EndFor
\State $\bar{x} \gets x^{(T)}$
\end{algorithmic}
\end{algorithm*}

\section{$L_1$ Attack} \label{sec:fw-pseudo}

We chose to use the Frank-Wolfe algorithm for optimizing the $L_1$ attack, 
as Projected Gradient Descent would require projecting onto a truncated $L_1$ ball, 
which is a complicated operation. In contrast, Frank-Wolfe only requires 
optimizing linear functions $g^{\top}x$ over a truncated $L_1$ ball; this can be done by 
sorting coordinates by the magnitude of $g$ and moving the top $k$ coordinates to the 
boundary of their range (with $k$ chosen by binary search). This is detailed in Algorithm
\ref{alg:fw-alg}.

\end{document}